%% file: main.tex
\documentclass[10pt,twocolumn,letterpaper]{article}

\usepackage[pagenumbers]{iccv}      % To produce the REVIEW version
\definecolor{iccvblue}{rgb}{0.21,0.49,0.74}
\usepackage[pagebackref,breaklinks,colorlinks,allcolors=iccvblue]{hyperref}

\usepackage{algorithm}
\usepackage{algorithmic}
\usepackage{amsmath,amssymb}
\usepackage{graphicx}
\usepackage{multirow}           
\usepackage{float}          
\usepackage{colortbl}      
\usepackage{caption}    
\usepackage{booktabs}
\usepackage{pifont}
\usepackage{array} 
\PassOptionsToPackage{table,xcdraw}{xcolor}
\usepackage{xcolor}

\newcommand{\xmark}{\ding{55}} 

\def\paperID{7676} % *** Enter the Paper ID here
\def\confName{ICCV}
\def\confYear{2025}

\title{DynCIM: Dynamic Curriculum for Imbalanced Multimodal Learning}
\author{
\hspace{-0.9cm}
\textbf{Chengxuan Qian}\textsuperscript{1}, 
\textbf{Kai Han}\textsuperscript{1}, 
\textbf{Jiaxin Liu}\textsuperscript{2}, 
\textbf{Zhenlong Yuan}\textsuperscript{3,4},
\textbf{Zhengzhong Zhu}\textsuperscript{5},\\
\textbf{Jingchao Wang}\textsuperscript{6},
\textbf{Chongwen Lyu}\textsuperscript{1},
\textbf{Jun Chen}\textsuperscript{1}, 
\textbf{Zhe Liu}\textsuperscript{1,$\dagger$}
\vspace{0.5em}
\\
{\small
\textsuperscript{1}Jiangsu University ~
\textsuperscript{2}UIUC ~
\textsuperscript{3}Alibaba  ~
\textsuperscript{4}UCAS ~
\textsuperscript{5}Sichuan University  ~
\textsuperscript{6}Peking University 
}\\
{\small
chengxuan.qian@stmail.ujs.edu.cn, zliu@ujs.edu.cn
}
\vspace{-1em}
}

\begin{document}
\maketitle
\input{sec/0_abstract}    
\input{sec/1_intro}

\begin{figure*}[ht]
    \centering
    \includegraphics[width=14cm]{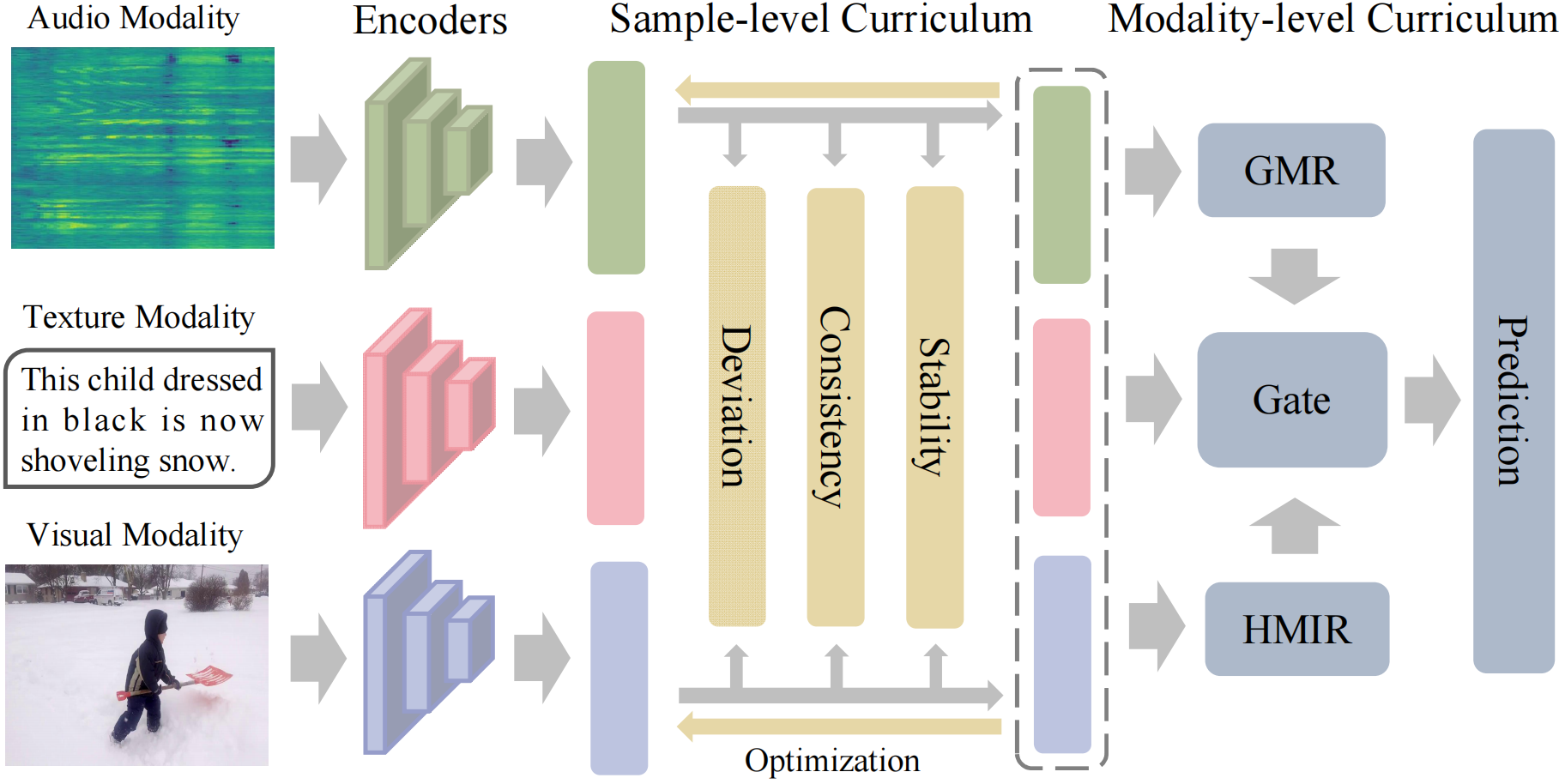}
    \caption{Overview of our proposed DynCIM curriculum learning framework. DynCIM leverages unimodal encoders to encode multimodal inputs, yielding outputs and byproducts of backpropagation. The Sample-level Curriculum accounts for sample-level imbalances by evaluating task-specific difficulty based on prediction deviation, consistency, and stability. Meanwhile, the Modality-level Curriculum adopts a global-to-local approach, introducing the Geometric Mean Ratio (GMR) to evaluate overall modality impact and the Harmonic Mean Improvement Ratio (HMIR) to capture fine-grained variations. A gating mechanism further regulates modality contributions, dynamically adjusting weights to mitigate modality imbalances. Finally, the refined multimodal representations are fused to generate the final prediction output, ensuring an adaptive and progressive learning process.}
    \label{pipeline}
    \vspace{-0.4cm}
\end{figure*}

\input{sec/2_related}

\input{sec/3_method}

\input{sec/4_experiments}

{
    \small
    \bibliographystyle{ieeenat_fullname}
    \bibliography{main}
}

\end{document}

%% file: sec/0_abstract.tex
\begin{abstract}
Multimodal learning integrates complementary information from diverse modalities to enhance the decision-making process. However, the potential of multimodal collaboration remains under-exploited due to disparities in data quality and modality representation capabilities. To address this, we introduce DynCIM, a novel dynamic curriculum learning framework designed to quantify the inherent imbalances from both sample and modality perspectives. DynCIM employs a sample-level curriculum to dynamically assess each sample's difficulty according to prediction deviation, consistency, and stability, while a modality-level curriculum measures modality contributions from global and local. Furthermore, a gating-based dynamic fusion mechanism is introduced to adaptively adjust modality contributions, minimizing redundancy and optimizing fusion effectiveness. Extensive experiments on six multimodal benchmarking datasets, spanning both bimodal and trimodal scenarios, demonstrate that DynCIM consistently outperforms state-of-the-art methods. Our approach effectively mitigates modality and sample imbalances while enhancing adaptability and robustness in multimodal learning tasks.
\vspace{-1em}
\end{abstract}

%% file: sec/1_intro.tex
\section{Introduction}
\label{sec:intro}

\begin{figure}[t!]
\centering
    \begin{subfigure}[t]{.23\textwidth}
        \centering
        \includegraphics[width=\textwidth]{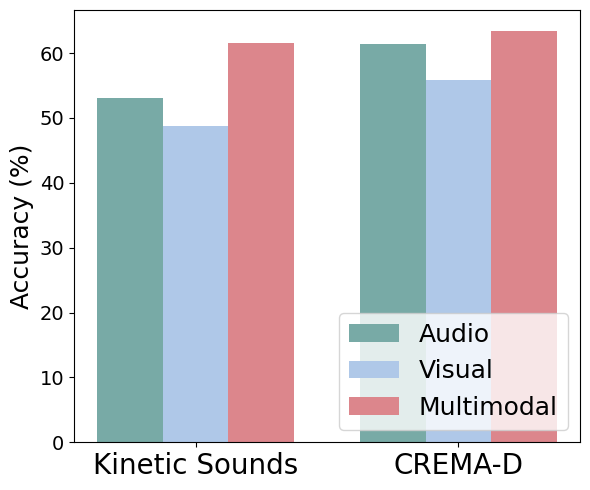}
        \caption{Modality performance gap}
        \label{fig:naive}
    \end{subfigure}
    \begin{subfigure}[t]{.23\textwidth}
        \centering
        \includegraphics[width=\textwidth]{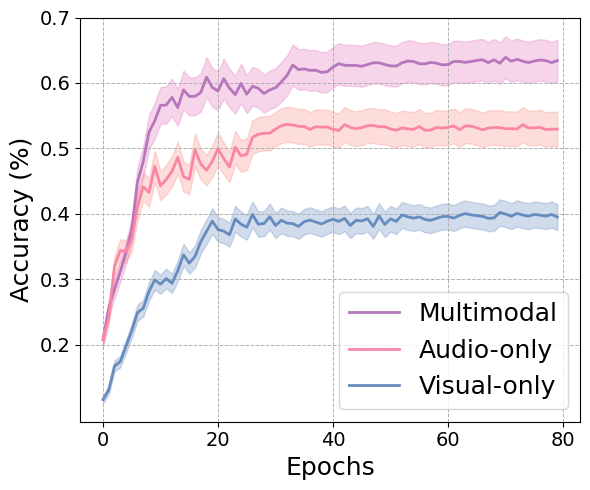}
        \caption{Multimodal convergence graph}
        \label{fig:reversed}
    \end{subfigure}
    \vspace{0.1cm}

    % 包含第三和第四幅图的容器
    \begin{subfigure}[t]{\columnwidth}
        \centering
        % 第三幅图
        \includegraphics[width=0.48\textwidth]{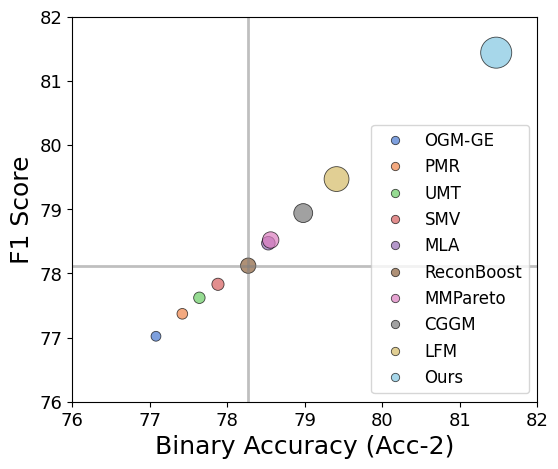}
        % 第四幅图
        \includegraphics[width=0.49\textwidth]{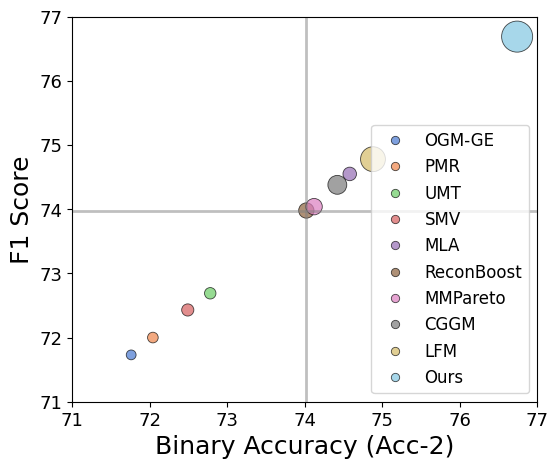}
        \caption{Superior Performance on Kinetics Sounds and CMU-MOSEI dataset}
        \label{fig:reversed_combined}
    \end{subfigure}

    \caption{(a) Performance gaps across modalities, indicating the dominance of certain modalities and the inherent limitations of naive fusion. (b) Variations in convergence rates on the Kinetics Sounds dataset, uncovering modality-specific learning dynamics. (c) Our approach consistently outperforms baselines on both bimodal and trimodal datasets, demonstrating its adaptability across different multimodal scenarios.}
    \label{fig:fig1}
    \vspace{-2em}
\end{figure}

% P1
Multimodal learning exploits and integrates heterogeneous yet complementary semantic information from various modalities to mimic human perception and decision-making \cite{zhang2024multimodal,xu2023mmsurvey,qian2025decalign,yuan2025autodrive,xing2025re,zhang2025pure}. However, the inherent heterogeneity across modalities poses challenges in mining cross-modal correlations and achieving effective inter-modal collaboration \cite{li2023dmd,jabeen2023review,liang2024foundations}. Existing studies indicate that in tasks like multimodal sentiment analysis \cite{meng2024deep,li2023dmd,das2023multimodal,yuan2025video}, action recognition \cite{shen2023towards,peng2022OGMGE,arandjelovic2017KS}, autonomous driving \cite{xing2024autotrust,xing2024openemma,hwang2024emma}, and large vision-language models \cite{xing2025re,wang2024enhancing,radford2021learning}, incorporating additional modalities often yield slight performance improvements over the strongest individual modality, as shown in Fig \ref{fig:fig1} (a). This suggests that the potential for inter-modal cooperation remains under-exploited, multimodal models still struggle with under-optimized challenges.

\begin{figure}[t]
    \centering
    \begin{subfigure}[t]{.45\linewidth}
        \centering
        \includegraphics[width=\textwidth, trim=0.05cm 0.08cm 0.08cm 0.08cm, clip]{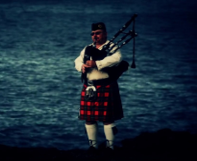}
    \end{subfigure}
    \hspace{4pt}
    \begin{subfigure}[t]{.45\linewidth}
        \centering
        \includegraphics[width=\textwidth, trim=0.05cm 0.08cm 0.08cm 0.08cm, clip]{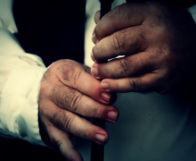}
    \end{subfigure}
    \caption{Different samples within the same modality exhibit varying levels of granularity, with the left image depicting the overall action category and contextual environment, while the right image highlights fine-grained movement details.}
    \vspace{-2em}
    \label{fig:fig2}
\end{figure}

% P2
To fully unlock the potential of multimodal models and address existing under-optimized challenges, we identify three key issues that hinder effective multimodal learning: \ding{182} \textbf{Inconsistent Learning Efficiency}, different modalities converge at varying rates, resulting in inconsistent learning efficiency \cite{wang2020makes,wei2024pamifly}, as shown in Fig. \ref{fig:fig1} (b). Therefore, ignoring modality-specific characteristics and applying uniform optimization strategies leads to suboptimal performance, necessitating a tailored approach that dynamically adapts to each modality’s convergence rate and representation capability. \ding{183} \textbf{Modality Imbalance}, the limited interpretability of deep models hinders the ability to observe the fine-grained contributions of each modality. Existing studies \cite{wu2022greedy,fan2023pmr,yang2024modalpref,li2023revisiting,zhang2024MLA} show that models tend to prioritize dominant modalities while under-utilizing weaker ones, exacerbating inter-modal imbalances and further complicating the implementation of effective optimization strategies. \ding{184} \textbf{Sample Quality Variability}, the quality of multimodal data fluctuates across instances due to environmental factors or sensor issues \cite{zhang2023QMF,zhang2023provable,zhang2024multimodal}. As shown in Fig \ref{fig:fig2}, the informativeness and representation capabilities of the visual modality differ across samples, resulting in varying contributions to multimodal predictions \cite{zhang2024multimodal,wei2024SMV,cao2024pdf,han2022mmdynamics}. However, mainstream methods typically batch multimodal samples randomly during iterative training, disregarding the inherent imbalances in sample difficulty, which can hinder model convergence and performance. Overall, single static optimization strategies are insufficient in addressing the dynamic fluctuations of sample quality and modality contributions, ultimately limiting the potential for effective multimodal collaboration and the exploitation of complementary inter-modal strengths.

To address these challenges, we propose DynCIM, a dynamic multimodal curriculum learning framework that adaptively optimizes cross-modal collaboration from both sample and modality perspectives. Specifically, \textbf{Sample-level Dynamic Curriculum} leverages back-propagation byproducts without extra neural modules to assess the real-time difficulty of multimodal samples. A dynamic weighting strategy prioritizes volatile metrics, allowing the model to emphasize critical aspects of sample evaluation under varying conditions. Meanwhile, \textbf{Modality-level Dynamic Curriculum} quantifies each modality’s contribution from global and local perspectives, using the Geometric Mean to evaluate overall performance improvement and the Harmonic Mean to detect redundant modalities. This dual-perspective approach ensures a more balanced fusion process by optimizing both collective impact and individual contributions, ultimately enhancing multimodal learning efficiency. To mitigate inter-modal imbalances, we introduce a \textbf{Modality Gating Mechanism} that dynamically adjusts each modality’s contribution, suppressing redundancy while amplifying informative signals. An \textbf{adaptive balance factor} further refines fusion by shifting the model’s focus between overall effectiveness and individual modality optimization, enhancing inter-modal coordination.

By integrating these mechanisms, DynCIM effectively addresses both sample and modality imbalances, dynamically guiding training to emphasize representative instances while progressively incorporating more challenging ones. This adaptive learning framework ensures a balanced, efficient, and robust multimodal learning process, ultimately improving inter-modal collaboration and fully leveraging complementary strengths for more effective decision-making. To validate its effectiveness and generalizability, we evaluate DynCIM across six widely used multimodal benchmarks in both bimodal and trimodal scenarios. Our contributions are threefold:
\begin{itemize}
    \item 
    \textbf{Dynamic Curricum Learning}. We propose DynCIM, a multimodal curriculum learning framework that adaptively addresses sample- and modality-level imbalances, ensuring a progressive and balanced learning process.
    
    \item 
    \textbf{Adaptive Optimization Strategy}.
    We introduce a dual-level curriculum, combining Sample-level Curriculum for difficulty-aware training and Modality-level Curriculum for optimal modality fusion, with a Modality Gating Mechanism dynamically balancing modality contributions for effective multimodal learning.

    \item 
    \textbf{Empirical Effectiveness}.
    Extensive experiments on six multimodal benchmarks demonstrate that DynCIM consistently outperforms state-of-the-art methods in both multimodal fusion and imbalanced multimodal learning tasks, confirming its robustness and generalizability.
\end{itemize}

% \begin{itemize}
%     \item We propose a Dynamic Curriculum Learning framework to address intrinsic imbalances from both sample and modality perspectives, adaptively accesses and prioritizes multimodal samples based on their instantaneous difficulty and modality collaboration effectiveness, ensuring a more balanced and progressive learning process.
%     \item To adaptively focus on key learning aspects across varying conditions, we introduce a Dynamic Weighting Strategy that prioritizes metrics based on volatility. Besides, a Modality Gating Mechanism adjusts each modality's contribution,  balancing between optimizing overall fusion effectiveness and fine-tuning individual modalities, thereby minimizing the impact of redundant modalities.
%     \item Extensive experiments  validate the effectiveness of our plug-and-play method, demonstrating improved performance in both multimodal fusion and imbalanced multimodal tasks across diverse domains.
% \end{itemize}

%% file: sec/2_related.tex
\vspace{-1em}
\section{Related Work}
\subsection{Imbalanced Multimodal Learning}

Multimodal learning enhances decision-making by leveraging the strengths of each modality. However, modality imbalance arises when certain modalities dominate due to uneven unimodal representational capabilities and training distribution bias \cite{das2023revisiting,wei2024pamifly,xu2023mmsurvey}. To mitigate this, methods like OGM-GE \cite{peng2022OGMGE}, PMR \cite{fan2023pmr}, and SMV \cite{wei2024SMV} adjust modality contributions through gradient modulation, prototype clustering, and Shapley-based resampling, respectively. While effective, these approaches primarily focus on modality contributions, overlooking sample-level imbalances caused by data quality variations or latent noise. Randomly batching samples can degrade performance and slow convergence, particularly when low-quality data dominates early training stages. To address this, we propose a curriculum learning strategy that dynamically reweights samples based on their difficulty and inter-modal collaboration effectiveness, ensuring a progressive learning process from simpler to more challenging examples.

\subsection{Curriculum Learning}
Curriculum learning \cite{bengio2009CL} facilitates a gradual transition from simple to complex tasks during model training, guided by two components: the Difficulty Measurer and the corresponding Training Scheduler. This easy-to-hard strategy helps guide and denoise the learning process, accelerating convergence and enhancing generalization capabilities \cite{wang2021CLsurveyPAMI}. CMCL \cite{liu2021aclCMCL} and I\textsuperscript{2}MCL \cite{zhou2023intra} assess difficulty using pre-trained models, but their reliance on such models increases inference costs and limits applicability in resource-constrained scenarios. To overcome this, we propose a dynamic curriculum learning method that leverages back-propagation byproducts to assess real-time sample difficulty without extra neural modules. By dynamically adjusting training priorities based on sample difficulty and modality collaboration effectiveness, our approach enhances learning efficiency, accelerates convergence, and improves multimodal performance.

% \subsection{Dynamic Multimodal Fusion}
% A key challenge in multimodal fusion is capturing cross-modal correlations while maximizing each modality’s strengths and minimizing redundancy \cite{cao2024pdf,han2022tmc,zhang2023provable,zhang2024multimodal}. However, most fusion methods are static in nature, treating all modalities equally without accounting for data quality variations or dynamic representation shifts. To address this, we propose a dynamic gating mechanism that adaptively regulates modality contributions in real-time, suppressing redundancy and enhancing informative signals. By introducing a balance factor that shifts focus between overall fusion optimization and fine-tuning individual contributions, our approach ensures a more adaptive and efficient fusion strategy, improving multimodal learning performance and robustness.

%% file: sec/3_method.tex
\section{Proposed DynCIM}

\textbf{Motivation and Overview.} Multimodal Learning faces two key challenges: (1) heterogeneous data quality and convergence rates across modalities, leading to inconsistent learning dynamics, (2) modality imbalance, where dominant modalities suppress weaker ones, limiting the potential for effective cross-modal collaboration. To address these, we introduce DyCIM, a dynamic curriculum learning framework that adaptively prioritizes samples and modalities based on their real-time difficulty and contribution. 
DyCIM consists of three components: \ding{182} Dynamic curriculum that evaluates the instantaneous difficulty of each multimodal sample using metrics from both sample and modality perspectives (Section \ref{sec:sec1}-\ref{sec:sec2}). \ding{183} Dynamic fusion strategy that adaptively integrates dynamically weighted samples and modality contributions, balancing between overall fusion effectiveness and individual modality fine-tuning (Section \ref{sec:sec3}). \ding{184} Overall multimodal curriculum learning framework that regulates the learning process across diverse multimodal samples, prioritizing informative samples and modalities while minimizing redundancy, thus enhancing training efficiency and robustness (Section \ref{sec:sec4}).

\subsection{Model Formulation}
For multimodal tasks, given the dataset $\mathcal{S}=\{x_i, y_i\}_{i=1}^N \in \mathcal{X} \times \mathcal{Y}$, where $N$ is the size of $\mathcal{S}$. Each sample $\mathbf{x}_i=\{x_i^{(m)}\}_{m=1}^\mathcal{|M|}$ has $\mathcal{|M|}$ modalities, and $y_i \in \mathbb{R}^K$ is the corresponding label, $K$ is the size of label space. To leverage information from multiple modalities and achieve effective multimodal cooperation, each modality employs a specific pre-trained unimodal encoder to extract features of respective modal data, denoted as $\phi^m(\cdot): \mathcal{X} \mapsto \mathcal{Y}$. For modality $m$, its encoder $\phi^m(\cdot)$ can be regarded as a probabilistic model that convert an observation $x^{(m)}$ to a Softmax-based predictive distribution $p^{(m)}(\mathbf{y} \mid x^{(m)})= [p_1^m, p_2^m,..., p_K^m]$.

\subsection{Sample-level Dynamic Curriculum}
\label{sec:sec1}
To systematically evaluate sample difficulty from multiple perspectives relevant to real-world multimodal tasks, we propose a task-oriented sample-level curriculum that quantifies sample difficulty \(\mathcal{D}_i\) based on prediction deviation, consistency, and stability. This formulation enables a dynamic assessment of each sample’s learning complexity relative to the model’s evolving capacity, ensuring a more adaptive and balanced training process.

\subsubsection{Prediction Deviation}
Conventional methods rely solely on multimodal loss as the supervisory signal, which tends to amplify dominant modalities and exacerbate inter-modal imbalances. To address this issue, we introduce both unimodal and multimodal losses, which leverage modality uncertainty to estimate predictive perplexity and reduce biases in multimodal joint training as \begin{equation}\label{eq1}
    \mathcal{D}_{\mathcal{L}}=\mathcal{L}_{concat}+\sum_{m=1}^{\mathcal{|M|}}\delta_m \cdot \mathcal{L}_m, 
\end{equation}
the weight $\{\delta_m\}_{m=1}^M\!\!=\!\!\exp(-U_m) / \sum_{m=1}^M \exp(-U_m)$ is dynamically adjusted according to the entropy of each modality’s prediction space, $U_m = -\sum_{k=1}^{K} p_k^m \log(p_k^m)$, indirectly capturing each unimodal encoder’s maturity and adaptively adjusting the supervisory signal accordingly.

\subsubsection{Prediction Consistency}
To ensure each unimodal encoder supports multimodal training effectively, it is crucial to monitor the imbalances in decision-making across modalities. By measuring consistency between unimodal predictions and the fused multimodal result, we can address inter-modal discrepancies, promoting balanced collaboration among modalities as 
\begin{equation}\label{eq2}
    \mathcal{D}_C=\frac{1}{\mathcal{|M|}}\sum_{m=1}^{\mathcal{M}}\|p^m-p^{f}\|^2,
\end{equation}
where $p^f$ is the fused multimodal prediction probability. The squared Euclidean distance quantifies the divergence between the $m$-th modality's prediction and the fused result, with larger values indicating weaker, possibly dominated contributions. This metric dynamically tracks each modality's impact, enabling real-time adjustments to balance influences and enhance fusion effectiveness.

\subsubsection{Prediction Stability}
In high-noise or ambiguous-feature scenarios, even when multimodal encoders achieve consistent decisions, unpredictable environmental factors can still impact each modality to varying degrees. By quantifying the confidence of each modality’s decisions in complex contexts, we gain further insight into robustness and generalization imbalances across modalities as
\begin{equation}\label{eq3}
    \mathcal{D}_S=-\log(p^{+}_k) |_{k=y} +\sum_{k\neq y}\log(p_k^{-}),
\end{equation}
where $p^{+}_k$ and $p^{-}_k$ are the Softmax-based predicted probabilities of the correct and incorrect classes, respectively. This metric rewards correct predictions while penalizing incorrect ones, providing insight into the model's reliability across varied multimodal contexts by highlighting the stability of predictions under challenging conditions.

\subsubsection{Adaptive Multi-metric Weighting Strategy}

In real-world multimodal scenarios, unpredictable environmental factors and inconsistent data collection can unevenly impact samples, making it challenging for a single metric to capture the complex supervisory needs required. To address this, we propose an adaptive multi-metric weighting strategy that dynamically shifts the model’s focus toward critical aspects by identifying metrics with higher volatility using standardization and exponential moving average (EMA). By assigning greater weight to these volatile metrics, the model prioritizes challenging samples, optimizing learning across diverse scenarios. Specifically, all metrics are first standardized via a sigmoid transformation as 
\begin{equation}\label{eq4}
\mathcal{D}'_j = \frac{1}{1 + \exp(-{s_j} \cdot \mathcal{D}_j)},
\end{equation}
with $s_j=-1$ for negative indicators (e.g., loss) and $s_j=1$ for positive ones (e.g., consistency, stability), ensuring that higher values indicate greater sample difficulty. To track and smooth fluctuations, we then compute the exponential moving average (EMA) of each metric's changes as 
\begin{equation}\label{eq5}
    \mathcal{E}_{j,t}=\gamma \cdot \mathcal{E}_{j,t-1}  +(1-\gamma) \cdot |\mathcal{D}'_{j,t}-\mathcal{D}'_{j,t-1}|,
\end{equation}
where $\mathcal{E}_{j,t}$ is the EMA of volatility for the $j$-th metric at training step $t$, and $\gamma$ is a smoothing factor that emphasizes recent data, providing a responsive measure of the model's current capabilities. Metric weights are then dynamically assigned based on the normalized EMA of volatility, giving higher weights to metrics with greater variability as 
\begin{equation}\label{eq6}
\varPsi_j = \frac{\mathcal{E}_{j, t}}{\sum_{j \in \{\mathcal{L},C,S\}} \mathcal{E}_{j, t}}.
\end{equation}
These weights integrate metrics into a composite task-oriented difficulty score, $\mathcal{D}_{\text{Task}}=\sum_{\substack{j \in \{\mathcal{L}, S, C\}}} \varPsi_j \cdot \mathcal{D}'_j$, enabling adaptive assessment of sample-level difficulty and guiding the model to prioritize the most critical aspects of the multimodal task as training progresses.

% Leveraging the diverse strengths of multiple modalities is crucial for improving model performance in multimodal tasks. However, their contributions are often imbalanced, leading to inefficiencies when certain modalities fail to provide the expected boost upon fusion. To optimize the fusion strategy, we introduce two metrics: the Geometric Mean Ratio (GMR) and the Harmonic Mean Improvement Rate (HMIR), designed to evaluate each modality's fine-grained contribution to the fusion model.
%%%%%%%%%%%%%%%%%%%%%%%%%%%%%%%%%%%%%%%%%%%
\subsection{Modality-level Dynamic Curriculum}
\label{sec:sec2}
Effectively leveraging multiple modalities is key to enhancing multimodal learning, yet modeling individual modality contributions remains challenging, often leading to imbalances and inefficiencies. To address this, we introduce Geometric Mean Ratio (GMR) and Harmonic Mean Improvement Rate (HMIR), which assess modality contributions from global and local perspectives, enabling finer-grained quantification for balanced fusion.

\subsubsection{Geometric Mean Ratio}
The Geometric Mean Ratio (GMR) offers a global perspective on the fusion model's performance by capturing the overall improvement achieved through multimodal integration compared to individual modalities, thus highlighting the collective impact of fusion. GMR is defined as
\begin{equation}\label{eq7}
    \mathcal{D}_G = \left(\prod_{m=1}^{|\mathcal{M}|} \frac{\mathcal{L}_{concat}}{\mathcal{L}_{m}}\right)^{\frac{1}{|\mathcal{M}|}}.
\end{equation}
When the GMR significantly exceeds 1, it indicates a substantial overall improvement in model performance resulting from effective fusion. Conversely, GMR close to or below 1 suggests that certain modalities contribute redundantly or insignificantly, providing insight into the global efficiency of the fusion process.

\subsubsection{Harmonic Mean Improvement Rate}

In contrast to GMR's global focus, the Harmonic Mean Improvement Rate (HMIR) offers a local view of each modality’s fine-grained contribution, emphasizing the minimum impactful ones. This approach allows dynamic identification and adjustment of under-performing or redundant modalities in imbalanced tasks, ensuring no single modality disproportionately detracts from performance as

\begin{equation}\label{eq8}
\mathcal{D}_{\mathcal{H}} = \frac{|\mathcal{M}|}{\sum_{m=1}^{|\mathcal{M}|} \omega_m \left( 1+ \frac{1}{\text{Gain}_m + \epsilon} \right)}, 
\end{equation}
where $\epsilon=1e-6$ is a small positive constant to prevent division by zero, and $\text{Gain}_m$ indicates the performance gain brought by multimodal fusion over each individual unimodal network as 
\begin{equation}
\label{equ:gain}
    \text{Gain}_m= \exp \left(\frac{\mathcal{L}_m-\mathcal{L}_{concat}}{\mathcal{L}_m} \right),
\end{equation}
the weight $\omega_m$ reflects the relative importance of each modality based on its contribution during multimodal fusion and satisfies $\sum_{m=1}^{\mathcal{|M|}}\omega_m=1$. It is initialized as
\begin{equation}
\label{eq:dynweight}
    \omega_m=\frac{\exp(-\mathcal{L}_m)}{\sum_{m=1}^{\mathcal{|M|}}\exp(-\mathcal{L}_m)}.
\end{equation}
By focusing on each modality’s minimal contribution, HMIR enables precise, targeted adjustments, ensuring the model dynamically adapts to minimize the impact of under-performing modalities.

% As the model's capabilities continuously improve throughout iterative training, the weights can be dynamically updated to reflect each modality's current impact:
% \begin{equation}
%     \varPhi^*_m=\frac{\varPhi_m \times \text{Gain}_m}{\sum_{j=1}^{\mathcal{|M|}} \left( \varPhi_j \times \text{Gain}_j \right)}
% \end{equation}
% This dynamic updating weights enhance the fusion model's efficiency by ensuring that only the most relevant information from each modality is utilized, aligning the model’s performance closely with the actual utility of each modality.

\subsection{Dynamic Multimodal Fusion}
\label{sec:sec3}
To leverage the most informative modalities while minimizing the impact of redundant ones, we introduced a dynamic network architecture that adaptively suppresses redundant modalities and adjusts the weights of each modality throughout the training process.

\subsubsection{Modality Gating Mechanism}
The modality gating mechanism dynamically controls the activation or suppression of each modality based on the fusion-level dynamic curriculum, thereby determining each modality's contribution to decision-making. The gating function $g_m$ for the $m$-th modality is
\begin{equation}\label{eq11}
    g_m=\sigma \left( \lambda_m \mathcal{D}_G + \left(1-\lambda_m \right) \mathcal{D}_\mathcal{H} \right),
\end{equation}
where $\sigma$ is the sigmoid activation function, and $\lambda$ is a balancing factor that dynamically shifts the model's focus between overall fusion effectiveness and individual modality contributions. This balancing factor is defined as
\begin{equation}\label{eq12}
    \lambda_m = \lambda_0 + \frac{1}{2} \sigma \left( \left| \text{Gain}_m - \overline{\text{Gain}} \right| \right),
\end{equation}
where $\lambda_0$ is initialized to 0.5 to balance overall and individual contributions, while $\overline{\text{Gain}}=\frac{1}{\mathcal{|M|}}\sum_{m=1}^{\mathcal{|M|}}\text{Gain}_m$ represents the average gain across all modalities, helping adjust the relative importance of each modality based on its deviation from this average.
% , and $\beta$ is a hyper-parameter controlling the sensitivity to variations in the gains of different modalities.
\subsubsection{Multimodal Fusion and Prediction}
% Following the determination of activation states, the weights of active modalities are dynamically adjusted according to the modality-level curriculum:
% \begin{equation}\label{eq16}
%     \omega_m^*=\omega_m \times \left( 1 + g_m \right)
% \end{equation}
% where $\omega_m^*$ is the updated weight for the $m$-th modality, initially set based on Equation (\ref{eq:dynweight}), with $\mu$ as the learning rate controlling the magnitude of the adjustment.

Following the determination of activation states, the weights of active modalities are dynamically adjusted according to the modality-level curriculum, where the updated weight for the $m$-th modality is given by $\omega_m^*=\omega_m \times \left( 1 + g_m \right)$. Here, $\omega_m^*$ is initialized based on Equation (\ref{eq:dynweight}), with $\mu$ as the learning rate controls the adjustment magnitude. To aggregate the outputs from active modalities, the dynamically adjusted weights $\omega_m^*$ are applied to the respective modality features as
\begin{equation}\label{eq13}
\hat{y}_{multi}= \sum_{m \in \text{A}} \omega_m^* \cdot \phi^m (x^{(m)}),
\end{equation}
where $m \in A$ denotes the  the modalities activated by the gating mechanism, and $\phi^m (x^{(m)})$ is the unimodal output of modality $m$, ensuring that the most informative modalities contribute proportionally to the final decision and enhance overall multimodal integration effectiveness.

% $\hat{y}_m=\phi^m (x^{(m)})$

\subsection{Multimodal Curriculum Learning}
\label{sec:sec4}
To address the inherent cross-modal heterogeneity and varying quality across multimodal samples, we propose an integrated curriculum learning strategy that combines sample-level and modality-level dynamic curricula. This approach adaptively prioritizes samples and modalities based on their estimated difficulty and information richness, ensuring a structured and progressive learning process. Our approach dynamically adjusts sample importance according to its contribution to model progression. Specifically, the optimization objective is defined as:
\begin{equation}\label{eq14}
    \min_{\mathcal{V}} \sum_{i=1}^{N}  \underbrace{ v_i \left( \mathcal{D}_{\text{Task}, i} \cdot \mathcal{D}_{\text{Fuse},i} \right)}_{Reweighting} \cdot \mathcal{L}_{\text{Task}}
    + \underbrace{\eta(t) \left(1 - v_i\right)^2}_{Regularizer},
\end{equation} 
where $\mathcal{D}_{\text{Task}}=\sum_{\substack{j \in \{\mathcal{L}, S, C\}}} \varPsi_j \cdot \mathcal{D}'_j$ represents the task-oriented sample difficulty derived from prediction deviation, consistency, and stability, while $\mathcal{D}_{\text{fuse},i}=\frac{1}{{\mathcal{|M|}}} {\sum_{m=1}^{\mathcal{|M|}}g_{m,i}}$ reflects the effectiveness of multimodal fusion for the $i$-th sample. The reweighting factor $\mathcal{W}_i=\left( \mathcal{D}_{\text{task}, i} \cdot \mathcal{D}_{\text{fuse},i} \right)$ ensures the model prioritizes samples that are both informative and challenging, directing learning focus toward key aspects of multimodal optimization. To prevent overfitting or instability in sample weighting, we introduce a regularization term $\eta(t) \left(1 - v_i\right)^2$, which encourages smooth weight transitions and prevents the model from over-relying on specific samples. By jointly optimizing sample-level and modality-level curriculum, this strategy enables the model to dynamically adapt to data heterogeneity, effectively balance learning across modalities, and enhance robustness and efficiency in multimodal learning.

%% file: sec/4_experiments.tex
\section{Experiments}

%% Dataset
\noindent\textbf{Datasets.} We conduct experiments on six widely used multimodal benchmarks datasets, covering both bi-modal and tri-modal scenarios. \textbf{Kinetic Sounds (KS)} \cite{arandjelovic2017KS} is an audio-visual action recognition dataset across 31 human action classes from YouTube videos. \textbf{CREMA-D} \cite{cao2014crema} is an audio-visual dataset for speech emotion recognition, containing six emotions labeled by 2,443 crowd-sourced raters. \textbf{UCF-101} \cite{soomro2012ucf101} is a bi-modal action recognition dataset featuring both RGB and Optical Flow data across 101 human actions categories. \textbf{CMU-MOSI} \cite{zadeh2016multimodal}, \textbf{CMU-MOSEI} \cite{zadeh2018mosei} and \textbf{CH-SIMS} \cite{yu2020ch} are multimodal sentiment analysis datasets incorporating audio, visual, and textual modalities.

% \subsection{Experimental settings}
% Unless specified, ResNet-18 is used as the encoder. For text, we use pre-trained BERT for embeddings. All models are trained with a batch size of 64 using the SGD optimizer with 0.9 momentum and 1e-4 weight decay. The learning rate starts at 1e-3 and decays to 1e-4. The proposed framework all combine different modalities by adopting summation. The experiment went through 80 epochs. Experiments were conducted on an NVIDIA RTX A6000 GPU. Besides, accuracy is adopted to evaluate all methods. Additional details are provided in the \textit{Supplementary Materials}.

\noindent\textbf{Experimental settings.} All experiments are conducted on an NVIDIA A6000 Ada with 48GB memory. Specifically, for the KS, CREMA-D and UCF-101 datasets, we adopt ResNet-18 as the backbone to ensure a fair comparison, following prior works \cite{wei2024pamifly,wei2024mmpareto, hua2024reconboost, zhang2024MLA}. For the MOSI, MOSEI, and CH-SIMS datasets, all datasets and experimental settings are implemented by MMSA-FET toolkit \cite{yu2021learning}.

\begin{table}[ht]
\centering
\footnotesize
\resizebox{0.47\textwidth}{!}{
\begin{tabular}{cccc}
\hline
\multirow{2}{*}{\textbf{Method}} & \textbf{KS} & \textbf{CREMA-D} & \textbf{UCF-101}\\
     & (Audio+Visual) & (Audio+Visual) & (RGB+OF) \\
\hline
\multicolumn{4}{l}{\cellcolor{gray!12} \textit{Unimodal Baselines}} \\
Audio/RGB-only      & 53.01 & 61.33 & 65.97 \\
Visual/OF-only      & 48.75 & 55.85 & 78.33 \\
\hline
\multicolumn{4}{l}{\cellcolor{gray!12} \textit{Conventional Multimodal Fusion Methods}} \\
Concatenation         & 61.57 & 63.47 & 81.18 \\
Summation      & 61.03 & 62.67 & 80.69  \\
FiLM \cite{perez2018film}       & 60.88 &  62.05 &  79.33 \\
BiGated \cite{kiela2018efficient}      & 62.41 & 63.88  & 81.50  \\
\hline
\multicolumn{4}{l}{\cellcolor{gray!12} \textit{Dynamic Multimodal Fusion Methods}} \\
TMC (TPAMI'22)       & 65.88 & 65.57  & 82.46 \\
QMF (ICML'23)         & 63.83 &  65.03 &  82.09 \\
PDF (ICML'24)        & 66.18 & 66.67  &  83.67 \\
\hline
\multicolumn{4}{l}{\cellcolor{gray!12} \textit{Imbalanced Multimodal Learning Methods}} \\
OGM-GE (CVPR'22)      & 63.48 & 65.58 & 82.45 \\
PMR (CVPR'23)      & 63.33 & 66.07 &  82.31 \\
I\textsuperscript{2}MCL (MM'23)    & 65.22 & 65.58 & 81.96 \\
UMT (CVPR'24)       & 66.08 & 66.68  &  82.62 \\
SMV (CVPR'24)       & 65.89 & 67.27 &  83.04 \\
MLA (CVPR'24)       & 67.55 &  69.53 &  84.21 \\
MMPareto (ICML'24)      & 67.18 & 71.08  &  84.89 \\
ReconBoost (ICML'24)      & 68.22 & 71.62  &  85.41 \\
CGGM (NeurIPS'24)      & 68.31 & 71.41  &  85.88 \\
LFM (NeurIPS'24)       & 68.77 &  72.88 &  86.01 \\
\hline
DynCIM (Ours) & \textbf{71.22} & \textbf{75.55} & \textbf{88.76} \\
\hline
\end{tabular}
}
\caption{Performance comparison on three bi-modal benchmark datasets: KS, CREMA-D, and UCF-101 for both multimodal fusion and imbalanced multimodal learning tasks.}
\vspace{-1em}
\label{tab:t1}
\end{table}

\begin{table}[ht]
\centering
\footnotesize
\resizebox{0.47\textwidth}{!}{
\begin{tabular}{cccc}
\hline
\multirow{1}{*}{\textbf{Method}} & \textbf{CMU-MOSI} & \textbf{CMU-MOSEI} & \textbf{CH-SIMS}\\
\hline
\multicolumn{4}{l}{\cellcolor{gray!12} \textit{Unimodal Baselines}} \\
Audio-only       & 54.69 & 52.21 & 58.17 \\
Visual-only      & 57.55 & 58.04 & 62.88 \\
Text-only        & 75.58 & 66.27 & 70.21 \\
\hline
\multicolumn{4}{l}{\cellcolor{gray!12} \textit{Conventional Multimodal Fusion Methods}} \\
Concatenation         & 76.57 & 67.78 & 73.31 \\
Summation      & 76.08 & 67.35 &  73.01 \\
FiLM \cite{perez2018film}       & 75.94 &  67.01 & 72.68  \\
BiGated \cite{kiela2018efficient}      & 76.97 & 68.01  & 73.78  \\
\hline
\multicolumn{4}{l}{\cellcolor{gray!12} \textit{Dynamic Multimodal Fusion Methods}} \\
TMC (TPAMI'22)       & 77.89 & 73.72  & 75.54 \\
QMF (ICML'23)         & 77.06 & 73.11  &  75.12 \\
PDF (ICML'24)        & 78.44 & 74.42 &  76.01 \\
\hline
\multicolumn{4}{l}{\cellcolor{gray!12} \textit{Imbalanced Multimodal Learning Methods}} \\
OGM-GE (CVPR'22)      & 77.08 & 71.76 & 74.03 \\
PMR (CVPR'23)      & 77.42 &  72.04 & 74.24 \\
I\textsuperscript{2}MCL (MM'23)    & 77.21 & 72.43 & 74.67 \\
UMT (CVPR'24)       & 77.64 &  72.78 &  75.35 \\
SMV (CVPR'24)       & 77.88 &  72.49 &  75.07 \\
MLA (CVPR'24)       & 78.53 &  74.58 &  77.42 \\
ReconBoost (ICML'24)      & 78.27 & 74.02  & 77.15  \\
MMPareto (ICML'24)      & 78.56 & 74.12  &  77.78 \\
CGGM (NeurIPS'24)      & 78.98 &  74.42 &  77.64 \\
LFM (NeurIPS'24)       & 79.41 &  74.88 &  78.13 \\

\hline
DynCIM (Ours) & \textbf{81.47} & \textbf{76.74} & \textbf{80.04} \\
\hline
\end{tabular}
}
\caption{Performance comparison on three tri-modal benchmark datasets: CMU-MOSI, CMU-MOSEI, and CH-SIMS, involving audio, visual and texture modalities for both multimodal fusion and imbalanced multimodal learning tasks.}
\vspace{-2em}
\label{tab:t2}
\end{table}

\subsection{Comparsion with State-of-the-Arts} 

\noindent\textbf{Conventional multimodal fusion methods}. Feature-level fusion methods like Concatenation, FiLM, and BiGated outperform the decision-level Summation approach, which, despite its interpretability, serves better as a foundational component within DynCIM. While Concatenation and BiGated enhance feature integration and boost performance, they lack adaptability to dynamic modality contributions, limiting their effectiveness in addressing sample and modality imbalances.

\noindent \textbf{Dynamic multimodal fusion methods}. To further explore the benefits of adaptive fusion, we compare DynCIM with TMC \cite{han2022tmc}, QMF \cite{zhang2023provable}, and PDF \cite{cao2024pdf}, which dynamically adjust modality weights to enhance performance. While these methods improve upon static fusion by leveraging uncertainty-based weighting, they are more susceptible to noise sensitivity due to reliance on uncertainty estimation. In contrast, DynCIM employs a gating mechanism that adaptively modulates each modality’s contribution based on real-time effectiveness, reducing redundancy and minimizing the impact of noisy modalities. Additionally, our adaptive balance factor dynamically shifts the focus between optimizing overall fusion and fine-tuning individual modality contributions, making DynCIM more robust and adaptable across diverse multimodal scenarios.

\noindent \textbf{Imbalanced multimodal learning methods}. To evaluate the effectiveness in addressing multimodal imbalances, we compare it againest OGM-GE \cite{peng2022balanced}, PMR \cite{fan2023pmr}, I\textsuperscript{2}MCL \cite{zhou2023intra}, UMT \cite{du2021improving}, SMV \cite{wei2024SMV}, MLA \cite{zhang2024MLA}, ReconBoost \cite{hua2024reconboost}, MMPareto \cite{wei2024mmpareto}, CGGM \cite{guo2024classifier} and LFM \cite{yang2024facilitating} in (Table \ref{tab:t1} \& \ref{tab:t2}). DynCIM consistently outperforms all methods in both bi-modal (KS, CREMA-D, UCF-101) and tri-modal (CMU-MOSI, CMU-MOSEI, CH-SIMS) settings. While MMPareto and ReconBoost improve modality alignment and CGGM and LFM refine fusion strategies, they rely on static optimization and struggle with adaptive adjustments. In contrast, DynCIM dynamically reweights samples and modalities through gradient-driven curriculum learning, prioritizing informative instances and balancing modality contributions in real-time. With a computational complexity of $O(N \cdot \mathcal{M})$, DynCIM achieves state-of-the-art performance, offering superior efficiency, adaptability, and robustness in multimodal learning.

% While SMV shares a similar motivation by considering sample-aware imbalances, its Shapley-based modality contribution estimation relies on static assessment and discrete optimization, lacking the fine-grained observation and dynamic adjustment of sample and modality contributions required for adaptive learning. Moreover, with a computational complexity of $ O (N \cdot 2^{|\mathcal{M}|} )$, SMV’s factorial complexity limits scalability as the number of modalities grows, making it computationally intensive for complex multimodal scenarios. In contrast, our method achieves efficiency by using gradient update byproducts to dynamically assess sample and modality importance, with no extra computational cost. Our dynamic curriculum reweights samples based on real-time difficulty and collaboration effectiveness, while the modality-level curriculum adaptively reduces low-quality modality impact. With a computational complexity of $O(N \cdot \mathcal{M})$, our approach scales effectively for large multimodal datasets, enhancing efficiency, robustness, and overall performance.

\begin{figure}[t]
    \centering
    \begin{subfigure}[t]{.45\linewidth}
        \centering
        \includegraphics[width=\textwidth, trim=0.05cm 0.08cm 0.08cm 0.08cm, clip]{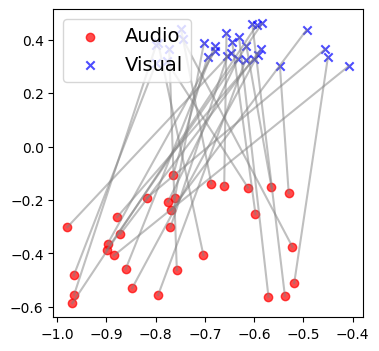}
        \caption{Concat Fusion}
        \label{gap:concat}
    \end{subfigure}
    \hspace{4pt}
    \begin{subfigure}[t]{.45\linewidth}
        \centering
        \includegraphics[width=\textwidth, trim=0.05cm 0.08cm 0.08cm 0.08cm, clip]{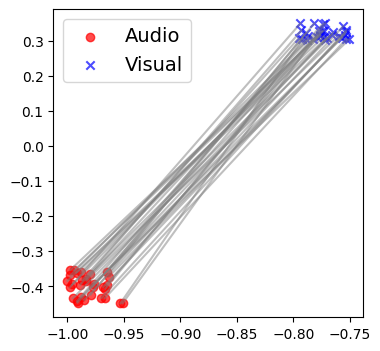}
        \caption{DynSIM (Ours)}
        \label{gap:ours}
    \end{subfigure}
    \vspace{-0.5em}
    \caption{Visualization of the modality gap between Audio and Visual on Kinetic Sounds dataset.}
    \vspace{-1em}
    \label{fig:gap}
\end{figure}

\noindent \textbf{Modality Gap Analysis}. Fig \ref{fig:gap} visualizes the modality gap between Audio and Visual features on the KS dataset, comparing Concat Fusion and DynCIM. In Concat Fusion, the features are scattered and poorly aligned, indicating weak cross-modal consistency and ineffective fusion. In contrast, DynCIM significantly reduces the modality gap, producing a tightly aligned feature distribution. This improvement stems from MDC’s dynamic modality contribution adjustment and gating mechanism, which together enhance inter-modal alignment by suppressing redundancy and amplifying informative signals, resulting in a more cohesive multimodal representation.

%%%% Ablation Table of SDC and MDC

\begin{table}[ht]
\centering
\footnotesize
\resizebox{\columnwidth}{!}{ 
\begin{tabular}{c@{\hskip 0.5cm}c|ccc}
\hline
\multicolumn{2}{c|}{\textbf{Module}} & \textbf{KS} & \textbf{CREMA-D} & \textbf{UCF-101}  \\
SDC & MDC & (Audio+Visual) & (Audio+Visual) & (RGB+OF) \\
\hline
\xmark & \xmark  & 62.19 & 65.23 & 81.69  \\
\checkmark & \xmark &  66.45 & 70.22  & 84.50  \\
 \xmark & \checkmark   & 68.19 & 72.73 & 86.29 \\
\checkmark & \checkmark  & \textbf{71.22} & \textbf{75.55} & \textbf{88.76} \\

\hline
\end{tabular}
}
\caption{Ablation study on Sample-level Dynamic Curriculum (SDC) and Modality-level Dynamic Curriculum (MDC).}
\vspace{-1em}
\label{tab:Ab1}
\end{table}

%%%% Ablation Table of Sample-level

\begin{table}[ht]
\centering
\footnotesize
\resizebox{\columnwidth}{!}{ 
\begin{tabular}{c@{\hskip 0.5cm}cc|ccc}
\hline
\multicolumn{3}{c|}{\textbf{Measure}} & \textbf{KS} & \textbf{CREMA-D} & \textbf{UCF-101} \\
$\mathcal{D}_{\mathcal{L}}$ & $\mathcal{D}_S$ & $\mathcal{D}_C$ & (Audio+Visual) & (Audio+Visual) & (RGB+OF) \\
\hline
\xmark & \xmark  & \xmark   & 65.19  & 67.73 & 81.29 \\
\checkmark & \xmark &  \xmark  &  68.70 & 71.80  & 84.94 \\
\checkmark & \checkmark   & \xmark & 70.03 & 74.12 & 87.05 \\
\checkmark & \checkmark  & \checkmark & \textbf{71.22} & \textbf{75.55} & \textbf{88.76} \\
\hline
\end{tabular}
}
\caption{Ablation study on different metrics for Sample-level Dynamic Curriculum (SDC).}
\label{tab:Ab2}
\vspace{-1em}
\end{table}

%%%% Ablation of Modality-level

\begin{table}[ht]
\centering
\footnotesize
\resizebox{\columnwidth}{!}{ 
\begin{tabular}{c@{\hskip 0.5cm}c|ccc}
\hline
\multicolumn{2}{c|}{\textbf{Measure}} & \textbf{KS} & \textbf{CREMA-D} & \textbf{UCF-101} \\
GMR $\mathcal{D}_G$  & HMIR $\mathcal{D}_{\mathcal{H}}$  & (Audio+Visual) & (Audio+Visual) & (RGB+OF) \\
\hline
\xmark & \xmark  & 66.45  & 69.22 & 82.50 \\
\checkmark & \xmark    & 69.86 & 72.81 & 85.34 \\
\xmark & \checkmark     & 69.37 & 73.55 & 86.07 \\
\checkmark & \checkmark   & \textbf{71.22} & \textbf{75.55} & \textbf{88.76} \\
\hline
\end{tabular}
}
\caption{Ablation study on different metrics for Modality-level Dynamic Curriculum (MDC)}
\label{tab:Ab3}
\vspace{-1em}
\end{table}

%%%% Ablation of gate

\begin{table}[ht]
\centering
\footnotesize
\begin{tabular}{@{\hskip 0.5cm}c|ccc}
\hline
Methods & \textbf{KS} &\textbf{CREMA-D}& \textbf{UCF-101} \\
\hline
 Uniform Weighting &67.09 &71.35 & 84.18 \\
 Modality Gating Mechanism & \textbf{71.22}  & \textbf{75.55}& \textbf{88.76} \\
\hline
\end{tabular}
\caption{Ablation Study on Modality Gating Mechanism.}
\label{tab:Ab4}
\vspace{-1em}
\end{table}

\begin{figure*}[t]
\centering
    \begin{subfigure}[t]{.23\textwidth}
			\centering
			\includegraphics[width=\textwidth]{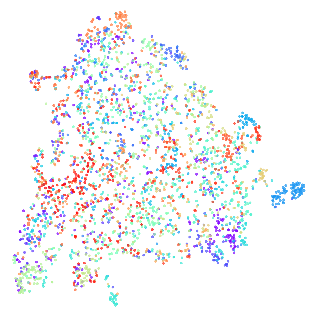}
			\caption{Concat Fusion}
			\label{fig:tsne-baseline}
	\end{subfigure}
	    \begin{subfigure}[t]{.23\textwidth}
			\centering
			\includegraphics[width=\textwidth]{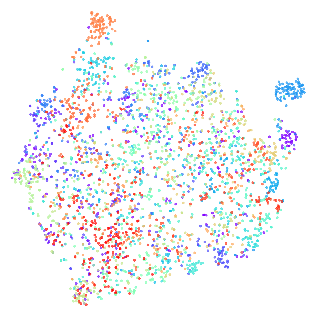}
			\caption{Ours w/o SDC}
			\label{fig:tsne-mmtm}
	\end{subfigure}
	\begin{subfigure}[t]{.23\textwidth}
			\centering
			\includegraphics[width=\textwidth]{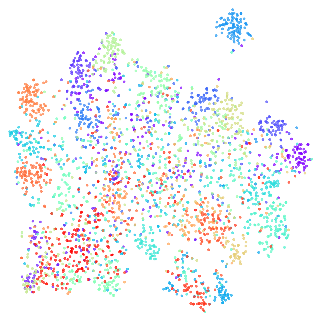}
			\caption{Ours w/o MDC}
			\label{fig:tsne-sample}
	\end{subfigure}
	\begin{subfigure}[t]{.23\textwidth}
			\centering
			\includegraphics[width=\textwidth]{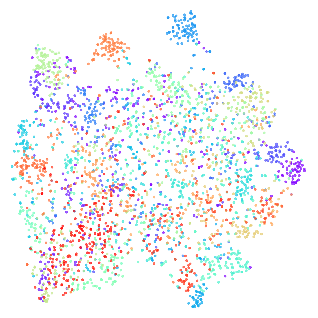}
			\caption{DynCIM (Ours)}
			\label{fig:tsne-modality}
	\end{subfigure}
 \vspace{-0.5em}
    \caption{t-SNE visualization \cite{van2008tsne} of feature distributions on KS dataset, comparing Concatenation, the method without SDC or MDC, and the complete method, with categories shown in different colors.}
        \vspace{-1.5em}
    \label{fig:tsne}
\end{figure*}

\subsection{Ablation Studies}
To validate the effectiveness of each component in DynCIM, we conducted a series of four ablation studies on the KS, CREMA-D, and UCF-101 datasets, systematically analyzing the contributions of sample-level (SDC) and modality-level (MDC) dynamic curriculum, their individual evaluation metrics, and the modality gating mechanism.

\noindent\textbf{Impact of SDC and MDC}. To assess the impact of SDC and MDC, we analyze quantitative results (Table \ref{tab:Ab1}) across three datasets and t-SNE visualizations (Fig \ref{fig:tsne}) on the KS dataset. Both components significantly enhance multimodal learning, with MDC providing greater improvement by dynamically optimizing modality contributions. The t-SNE results further support this: without SDC, feature distributions show weaker intra-class compactness, highlighting its role in refining sample representations and addressing imbalances. Without MDC, class separation is less distinct, demonstrating its importance in enhancing cross-modal integration. In contrast, the full DynCIM framework yields well-structured, compact feature distributions, confirming that SDC strengthens intra-class consistency while MDC optimizes inter-modal fusion, leading to more robust and discriminative multimodal representations.

\noindent\textbf{Impact of specific mechanisms}. We conduct three ablation studies to analyze DynCIM’s key mechanisms. First, evaluating individual SDC metrics (Table \ref{tab:Ab2}) shows that incorporating prediction deviation, consistency, and stability enhances performance, emphasizing the need for comprehensive difficulty assessment. Second, Table \ref{tab:Ab3} reveals that combining Geometric Mean Ratio (GMR) and Harmonic Mean Improvement Ratio (HMIR) achieves optimal results, balancing overall modality impact and fine-grained adjustments. Lastly, comparing the Modality Gating Mechanism with Uniform Weighting (Table \ref{tab:Ab4}) highlights its effectiveness in dynamically adjusting modality contributions, reducing redundancy, and improving robustness.

\begin{figure}[t]
\centering
    \begin{subfigure}[t]{.23\textwidth}
			\centering
			\includegraphics[width=\textwidth, trim=0.05cm 0.08cm 0.08cm 0.08cm, clip]{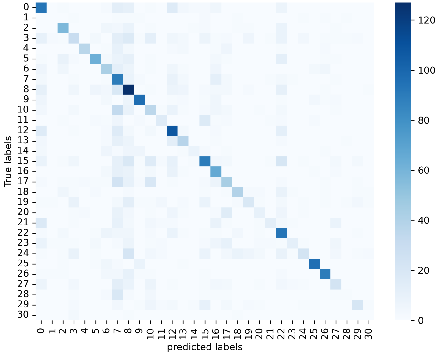}
			\caption{Concatenation-based Fusion}
			\label{c1}
	\end{subfigure}
	    \begin{subfigure}[t]{.23\textwidth}
			\centering
			\includegraphics[width=\textwidth, trim=0.05cm 0.08cm 0.08cm 0.08cm, clip]{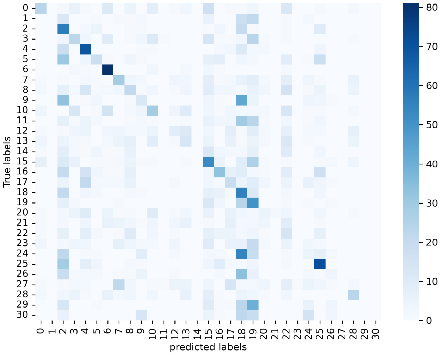}
			\caption{Only hard-to-easy training}
			\label{c2}
	\end{subfigure}
	\begin{subfigure}[t]{.23\textwidth}
			\centering
			\includegraphics[width=\textwidth, trim=0.05cm 0.08cm 0.08cm 0.08cm, clip]{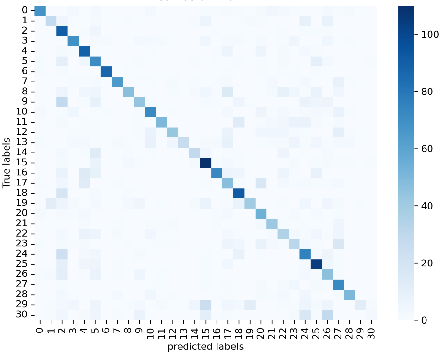}
			\caption{Only easy-to-hard training}
			\label{c3}
	\end{subfigure}
	\begin{subfigure}[t]{.23\textwidth}
			\centering
			\includegraphics[width=\textwidth, trim=0.05cm 0.08cm 0.08cm 0.08cm, clip]{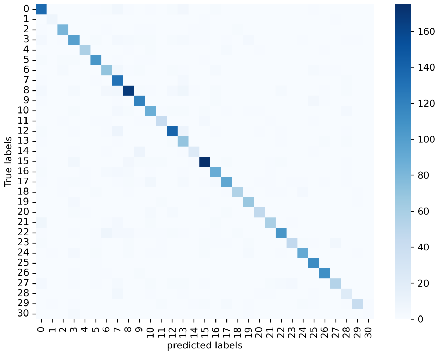}
			\caption{DynCIM (Ours)}
			\label{c4}
	\end{subfigure}
 \vspace{-0.5em}
    \caption{Visualization of ablation studies on different learning strategies on KS datasets}
        \vspace{-1.5em}
    \label{fig:confuse}
\end{figure}

\noindent\textbf{Impact of learning strategies}. To analyze the effectiveness of different learning strategies, we visualize their impact on the KS dataset in Fig \ref{fig:confuse}. Concatenation-based Fusion lacks a structured training approach, leading to higher mis-classification rates. Hard-to-easy training struggles with early-stage instability, resulting in scattered predictions and poor convergence. Easy-to-hard training shows improved alignment but lacks adaptability to sample variations. In contrast, DynCIM achieves the most structured and well-aligned predictions, demonstrating that our progressive curriculum learning strategy effectively balances stability and adaptability, leading to smoother convergence and improved accuracy.

\subsection{Parameter Sensitivity Analysis} 

\begin{figure}[ht]
    \centering
    \includegraphics[width=0.95\columnwidth, clip, trim=0cm 0cm 0cm 0cm]{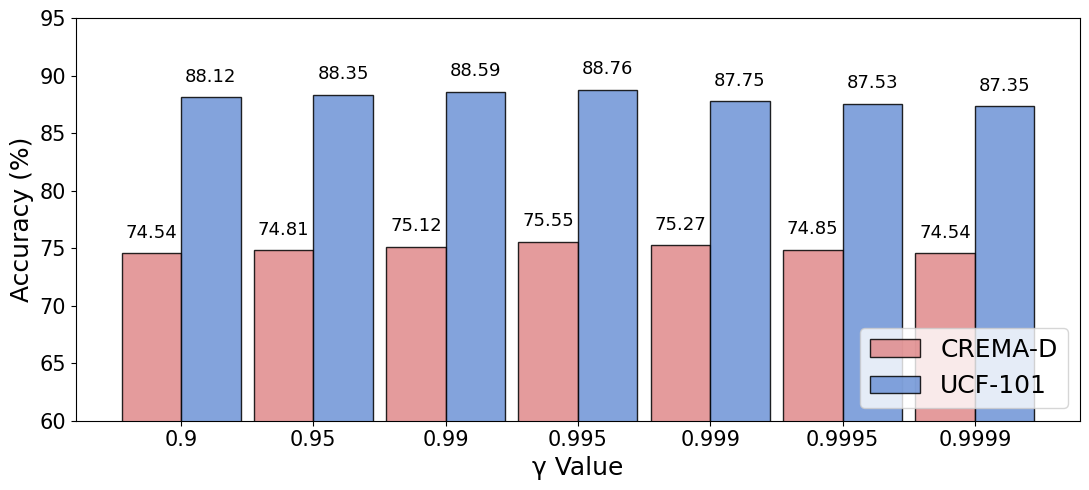}
    \vspace{-1em}
    \caption{Impact of the parameter $\gamma$ in EMA on accuracy performance for the CREAM-D and UCF-101 datasets, respectively.}\label{fig2}
    \vspace{-2em}
\end{figure}

The parameter $\gamma$ in the EMA update significantly influences sample-level metric performance in model fusion, affecting overall convergence. As shown in Figure \ref{fig2}, we tested various $\gamma$ values on the CREMA-D and UCF-101 datasets, with optimal accuracy achieved at $\gamma = 0.995$. Smaller $\gamma$ values cause current metrics to lag behind historical updates, hindering convergence. Conversely, larger $\gamma$ values reduce the discrepancy between current and historical metrics, limiting the historical average’s ability to correct biases in the current metric weighting.

\section{Conclusions}
We propose DynCIM, a dynamic multimodal curriculum learning framework that integrates sample and modality-level curriculum to adjust training priorities based on real-time difficulty and modality contributions. A gating mechanism further enhances fusion effectiveness, mitigating redundancy and improving adaptability. Extensive experiments on six benchmarks demonstrate DynCIM's superiority over state-of-the-art methods, achieving more balanced, robust, and efficient multimodal learning.

\clearpage

%% file: main.bbl
\begin{thebibliography}{53}
\providecommand{\natexlab}[1]{#1}
\providecommand{\url}[1]{\texttt{#1}}
\expandafter\ifx\csname urlstyle\endcsname\relax
  \providecommand{\doi}[1]{doi: #1}\else
  \providecommand{\doi}{doi: \begingroup \urlstyle{rm}\Url}\fi

\bibitem[Arandjelovic and Zisserman(2017)]{arandjelovic2017KS}
Relja Arandjelovic and Andrew Zisserman.
\newblock Look, listen and learn.
\newblock In \emph{Proceedings of the IEEE international conference on computer vision}, pages 609--617, 2017.

\bibitem[Bengio et~al.(2009)Bengio, Louradour, Collobert, and Weston]{bengio2009CL}
Yoshua Bengio, J{\'e}r{\^o}me Louradour, Ronan Collobert, and Jason Weston.
\newblock Curriculum learning.
\newblock In \emph{Proceedings of the 26th annual international conference on machine learning}, pages 41--48, 2009.

\bibitem[Cao et~al.(2024)Cao, Xia, Ding, Zhang, and Hu]{cao2024pdf}
Bing Cao, Yinan Xia, Yi Ding, Changqing Zhang, and Qinghua Hu.
\newblock Predictive dynamic fusion.
\newblock In \emph{International Conference on Machine Learning}. PMLR, 2024.

\bibitem[Cao et~al.(2014)Cao, Cooper, Keutmann, Gur, Nenkova, and Verma]{cao2014crema}
Houwei Cao, David~G Cooper, Michael~K Keutmann, Ruben~C Gur, Ani Nenkova, and Ragini Verma.
\newblock Crema-d: Crowd-sourced emotional multimodal actors dataset.
\newblock \emph{IEEE transactions on affective computing}, 5\penalty0 (4):\penalty0 377--390, 2014.

\bibitem[Das et~al.(2023)Das, Das, Sistu, Horgan, Bhattacharya, Jones, Glavin, and Eising]{das2023revisiting}
Arindam Das, Sudip Das, Ganesh Sistu, Jonathan Horgan, Ujjwal Bhattacharya, Edward Jones, Martin Glavin, and Ciar{\'a}n Eising.
\newblock Revisiting modality imbalance in multimodal pedestrian detection.
\newblock In \emph{2023 IEEE International Conference on Image Processing (ICIP)}, pages 1755--1759. IEEE, 2023.

\bibitem[Das and Singh(2023)]{das2023multimodal}
Ringki Das and Thoudam~Doren Singh.
\newblock Multimodal sentiment analysis: a survey of methods, trends, and challenges.
\newblock \emph{ACM Computing Surveys}, 55\penalty0 (13s):\penalty0 1--38, 2023.

\bibitem[Du et~al.(2021)Du, Li, Liu, Wen, Hua, Wang, and Zhao]{du2021improving}
Chenzhuang Du, Tingle Li, Yichen Liu, Zixin Wen, Tianyu Hua, Yue Wang, and Hang Zhao.
\newblock Improving multi-modal learning with uni-modal teachers.
\newblock \emph{arXiv preprint arXiv:2106.11059}, 2021.

\bibitem[Fan et~al.(2023)Fan, Xu, Wang, Wang, and Guo]{fan2023pmr}
Yunfeng Fan, Wenchao Xu, Haozhao Wang, Junxiao Wang, and Song Guo.
\newblock Pmr: Prototypical modal rebalance for multimodal learning.
\newblock In \emph{Proceedings of the IEEE/CVF Conference on Computer Vision and Pattern Recognition}, pages 20029--20038, 2023.

\bibitem[Guo et~al.(2024)Guo, Jin, Chen, and Zhao]{guo2024classifier}
Zirun Guo, Tao Jin, Jingyuan Chen, and Zhou Zhao.
\newblock Classifier-guided gradient modulation for enhanced multimodal learning.
\newblock \emph{Advances in Neural Information Processing Systems}, 37:\penalty0 133328--133344, 2024.

\bibitem[Han et~al.(2022{\natexlab{a}})Han, Yang, Huang, Zhang, and Yao]{han2022mmdynamics}
Zongbo Han, Fan Yang, Junzhou Huang, Changqing Zhang, and Jianhua Yao.
\newblock Multimodal dynamics: Dynamical fusion for trustworthy multimodal classification.
\newblock In \emph{Proceedings of the IEEE/CVF conference on computer vision and pattern recognition}, pages 20707--20717, 2022{\natexlab{a}}.

\bibitem[Han et~al.(2022{\natexlab{b}})Han, Zhang, Fu, and Zhou]{han2022tmc}
Zongbo Han, Changqing Zhang, Huazhu Fu, and Joey~Tianyi Zhou.
\newblock Trusted multi-view classification with dynamic evidential fusion.
\newblock \emph{IEEE transactions on pattern analysis and machine intelligence}, 45\penalty0 (2):\penalty0 2551--2566, 2022{\natexlab{b}}.

\bibitem[Hua et~al.(2024)Hua, Xu, Bao, Yang, and Huang]{hua2024reconboost}
Cong Hua, Qianqian Xu, Shilong Bao, Zhiyong Yang, and Qingming Huang.
\newblock Reconboost: Boosting can achieve modality reconcilement.
\newblock In \emph{International Conference on Machine Learning}, pages 19573--19597. PMLR, 2024.

\bibitem[Hwang et~al.(2024)Hwang, Xu, Lin, Hung, Ji, Choi, Huang, He, Covington, Sapp, et~al.]{hwang2024emma}
Jyh-Jing Hwang, Runsheng Xu, Hubert Lin, Wei-Chih Hung, Jingwei Ji, Kristy Choi, Di Huang, Tong He, Paul Covington, Benjamin Sapp, et~al.
\newblock Emma: End-to-end multimodal model for autonomous driving.
\newblock \emph{arXiv preprint arXiv:2410.23262}, 2024.

\bibitem[Jabeen et~al.(2023)Jabeen, Li, Amin, Bourahla, Li, and Jabbar]{jabeen2023review}
Summaira Jabeen, Xi Li, Muhammad~Shoib Amin, Omar Bourahla, Songyuan Li, and Abdul Jabbar.
\newblock A review on methods and applications in multimodal deep learning.
\newblock \emph{ACM Transactions on Multimedia Computing, Communications and Applications}, 19\penalty0 (2s):\penalty0 1--41, 2023.

\bibitem[Kiela et~al.(2018)Kiela, Grave, Joulin, and Mikolov]{kiela2018efficient}
Douwe Kiela, Edouard Grave, Armand Joulin, and Tomas Mikolov.
\newblock Efficient large-scale multi-modal classification.
\newblock In \emph{Proceedings of the AAAI conference on artificial intelligence}, 2018.

\bibitem[Li et~al.(2023{\natexlab{a}})Li, Fei, Liao, Zhao, Teng, Chua, Ji, and Li]{li2023revisiting}
Bobo Li, Hao Fei, Lizi Liao, Yu Zhao, Chong Teng, Tat-Seng Chua, Donghong Ji, and Fei Li.
\newblock Revisiting disentanglement and fusion on modality and context in conversational multimodal emotion recognition.
\newblock In \emph{Proceedings of the 31st ACM International Conference on Multimedia}, pages 5923--5934, 2023{\natexlab{a}}.

\bibitem[Li et~al.(2023{\natexlab{b}})Li, Wang, and Cui]{li2023dmd}
Yong Li, Yuanzhi Wang, and Zhen Cui.
\newblock Decoupled multimodal distilling for emotion recognition.
\newblock In \emph{Proceedings of the IEEE/CVF Conference on Computer Vision and Pattern Recognition}, pages 6631--6640, 2023{\natexlab{b}}.

\bibitem[Liang et~al.(2024)Liang, Zadeh, and Morency]{liang2024foundations}
Paul~Pu Liang, Amir Zadeh, and Louis-Philippe Morency.
\newblock Foundations \& trends in multimodal machine learning: Principles, challenges, and open questions.
\newblock \emph{ACM Computing Surveys}, 56\penalty0 (10):\penalty0 1--42, 2024.

\bibitem[Liu et~al.(2021)Liu, Ge, and Wu]{liu2021aclCMCL}
Fenglin Liu, Shen Ge, and Xian Wu.
\newblock Competence-based multimodal curriculum learning for medical report generation.
\newblock In \emph{Proceedings of the 59th Annual Meeting of the Association for Computational Linguistics and the 11th International Joint Conference on Natural Language Processing (Volume 1: Long Papers)}, pages 3001--3012, 2021.

\bibitem[Meng et~al.(2024)Meng, Shou, Ai, Yin, and Li]{meng2024deep}
Tao Meng, Yuntao Shou, Wei Ai, Nan Yin, and Keqin Li.
\newblock Deep imbalanced learning for multimodal emotion recognition in conversations.
\newblock \emph{IEEE Transactions on Artificial Intelligence}, 2024.

\bibitem[Peng et~al.(2022{\natexlab{a}})Peng, Wei, Deng, Wang, and Hu]{peng2022OGMGE}
Xiaokang Peng, Yake Wei, Andong Deng, Dong Wang, and Di Hu.
\newblock Balanced multimodal learning via on-the-fly gradient modulation.
\newblock In \emph{Proceedings of the IEEE/CVF conference on computer vision and pattern recognition}, pages 8238--8247, 2022{\natexlab{a}}.

\bibitem[Peng et~al.(2022{\natexlab{b}})Peng, Wei, Deng, Wang, and Hu]{peng2022balanced}
Xiaokang Peng, Yake Wei, Andong Deng, Dong Wang, and Di Hu.
\newblock Balanced multimodal learning via on-the-fly gradient modulation.
\newblock In \emph{Proceedings of the IEEE/CVF conference on computer vision and pattern recognition}, pages 8238--8247, 2022{\natexlab{b}}.

\bibitem[Perez et~al.(2018)Perez, Strub, De~Vries, Dumoulin, and Courville]{perez2018film}
Ethan Perez, Florian Strub, Harm De~Vries, Vincent Dumoulin, and Aaron Courville.
\newblock Film: Visual reasoning with a general conditioning layer.
\newblock In \emph{Proceedings of the AAAI conference on artificial intelligence}, 2018.

\bibitem[Qian et~al.(2025)Qian, Xing, Li, Zhao, and Tu]{qian2025decalign}
Chengxuan Qian, Shuo Xing, Shawn Li, Yue Zhao, and Zhengzhong Tu.
\newblock Decalign: Hierarchical cross-modal alignment for decoupled multimodal representation learning.
\newblock \emph{arXiv preprint arXiv:2503.11892}, 2025.

\bibitem[Radford et~al.(2021)Radford, Kim, Hallacy, Ramesh, Goh, Agarwal, Sastry, Askell, Mishkin, Clark, et~al.]{radford2021learning}
Alec Radford, Jong~Wook Kim, Chris Hallacy, Aditya Ramesh, Gabriel Goh, Sandhini Agarwal, Girish Sastry, Amanda Askell, Pamela Mishkin, Jack Clark, et~al.
\newblock Learning transferable visual models from natural language supervision.
\newblock In \emph{International conference on machine learning}, pages 8748--8763. PMLR, 2021.

\bibitem[Shen et~al.(2023)Shen, Huang, Yin, Zou, Rajan, and See]{shen2023towards}
Meng Shen, Yizheng Huang, Jianxiong Yin, Heqing Zou, Deepu Rajan, and Simon See.
\newblock Towards balanced active learning for multimodal classification.
\newblock In \emph{Proceedings of the 31st ACM International Conference on Multimedia}, pages 3434--3445, 2023.

\bibitem[Soomro(2012)]{soomro2012ucf101}
K Soomro.
\newblock Ucf101: A dataset of 101 human actions classes from videos in the wild.
\newblock \emph{arXiv preprint arXiv:1212.0402}, 2012.

\bibitem[Van~der Maaten and Hinton(2008)]{van2008tsne}
Laurens Van~der Maaten and Geoffrey Hinton.
\newblock Visualizing data using t-sne.
\newblock \emph{Journal of machine learning research}, 9\penalty0 (11), 2008.

\bibitem[Wang et~al.(2020)Wang, Tran, and Feiszli]{wang2020makes}
Weiyao Wang, Du Tran, and Matt Feiszli.
\newblock What makes training multi-modal classification networks hard?
\newblock In \emph{Proceedings of the IEEE/CVF conference on computer vision and pattern recognition}, pages 12695--12705, 2020.

\bibitem[Wang et~al.(2021)Wang, Chen, and Zhu]{wang2021CLsurveyPAMI}
Xin Wang, Yudong Chen, and Wenwu Zhu.
\newblock A survey on curriculum learning.
\newblock \emph{IEEE transactions on pattern analysis and machine intelligence}, 44\penalty0 (9):\penalty0 4555--4576, 2021.

\bibitem[Wang et~al.(2024)Wang, Chen, Wang, Zhou, Zhou, Yao, Zhou, Goldstein, Bhatia, Huang, et~al.]{wang2024enhancing}
Xiyao Wang, Jiuhai Chen, Zhaoyang Wang, Yuhang Zhou, Yiyang Zhou, Huaxiu Yao, Tianyi Zhou, Tom Goldstein, Parminder Bhatia, Furong Huang, et~al.
\newblock Enhancing visual-language modality alignment in large vision language models via self-improvement.
\newblock \emph{arXiv preprint arXiv:2405.15973}, 2024.

\bibitem[Wei and Hu(2024)]{wei2024mmpareto}
Yake Wei and Di Hu.
\newblock Mmpareto: boosting multimodal learning with innocent unimodal assistance.
\newblock \emph{arXiv preprint arXiv:2405.17730}, 2024.

\bibitem[Wei et~al.(2024{\natexlab{a}})Wei, Feng, Wang, and Hu]{wei2024SMV}
Yake Wei, Ruoxuan Feng, Zihe Wang, and Di Hu.
\newblock Enhancing multimodal cooperation via sample-level modality valuation.
\newblock In \emph{Proceedings of the IEEE/CVF Conference on Computer Vision and Pattern Recognition}, pages 27338--27347, 2024{\natexlab{a}}.

\bibitem[Wei et~al.(2024{\natexlab{b}})Wei, Hu, Du, and Wen]{wei2024pamifly}
Yake Wei, Di Hu, Henghui Du, and Ji-Rong Wen.
\newblock On-the-fly modulation for balanced multimodal learning.
\newblock \emph{IEEE Transactions on Pattern Analysis and Machine Intelligence}, 2024{\natexlab{b}}.

\bibitem[Wu et~al.(2022)Wu, Jastrzebski, Cho, and Geras]{wu2022greedy}
Nan Wu, Stanislaw Jastrzebski, Kyunghyun Cho, and Krzysztof~J Geras.
\newblock Characterizing and overcoming the greedy nature of learning in multi-modal deep neural networks.
\newblock In \emph{International Conference on Machine Learning}, pages 24043--24055. PMLR, 2022.

\bibitem[Xing et~al.(2024{\natexlab{a}})Xing, Hua, Gao, Zhu, Li, Tian, Li, Huang, Yang, Wang, et~al.]{xing2024autotrust}
Shuo Xing, Hongyuan Hua, Xiangbo Gao, Shenzhe Zhu, Renjie Li, Kexin Tian, Xiaopeng Li, Heng Huang, Tianbao Yang, Zhangyang Wang, et~al.
\newblock Autotrust: Benchmarking trustworthiness in large vision language models for autonomous driving.
\newblock \emph{arXiv preprint arXiv:2412.15206}, 2024{\natexlab{a}}.

\bibitem[Xing et~al.(2024{\natexlab{b}})Xing, Qian, Wang, Hua, Tian, Zhou, and Tu]{xing2024openemma}
Shuo Xing, Chengyuan Qian, Yuping Wang, Hongyuan Hua, Kexin Tian, Yang Zhou, and Zhengzhong Tu.
\newblock Openemma: Open-source multimodal model for end-to-end autonomous driving.
\newblock \emph{arXiv preprint arXiv:2412.15208}, 2024{\natexlab{b}}.

\bibitem[Xing et~al.(2025)Xing, Wang, Li, Bai, Wang, Qian, Yao, and Tu]{xing2025re}
Shuo Xing, Yuping Wang, Peiran Li, Ruizheng Bai, Yueqi Wang, Chengxuan Qian, Huaxiu Yao, and Zhengzhong Tu.
\newblock Re-align: Aligning vision language models via retrieval-augmented direct preference optimization.
\newblock \emph{arXiv preprint arXiv:2502.13146}, 2025.

\bibitem[Xu et~al.(2023)Xu, Zhu, and Clifton]{xu2023mmsurvey}
Peng Xu, Xiatian Zhu, and David~A Clifton.
\newblock Multimodal learning with transformers: A survey.
\newblock \emph{IEEE Transactions on Pattern Analysis and Machine Intelligence}, 2023.

\bibitem[Yang et~al.(2024{\natexlab{a}})Yang, Wan, Jiang, and Xu]{yang2024facilitating}
Yang Yang, Fengqiang Wan, Qing-Yuan Jiang, and Yi Xu.
\newblock Facilitating multimodal classification via dynamically learning modality gap.
\newblock \emph{Advances in Neural Information Processing Systems}, 37:\penalty0 62108--62122, 2024{\natexlab{a}}.

\bibitem[Yang et~al.(2024{\natexlab{b}})Yang, Wei, Liang, and Hu]{yang2024modalpref}
Zequn Yang, Yake Wei, Ce Liang, and Di Hu.
\newblock Quantifying and enhancing multi-modal robustness with modality preference.
\newblock In \emph{The Twelfth International Conference on Learning Representations}, 2024{\natexlab{b}}.

\bibitem[Yu et~al.(2020)Yu, Xu, Meng, Zhu, Ma, Wu, Zou, and Yang]{yu2020ch}
Wenmeng Yu, Hua Xu, Fanyang Meng, Yilin Zhu, Yixiao Ma, Jiele Wu, Jiyun Zou, and Kaicheng Yang.
\newblock Ch-sims: A chinese multimodal sentiment analysis dataset with fine-grained annotation of modality.
\newblock In \emph{Proceedings of the 58th annual meeting of the association for computational linguistics}, pages 3718--3727, 2020.

\bibitem[Yu et~al.(2021)Yu, Xu, Yuan, and Wu]{yu2021learning}
Wenmeng Yu, Hua Xu, Ziqi Yuan, and Jiele Wu.
\newblock Learning modality-specific representations with self-supervised multi-task learning for multimodal sentiment analysis.
\newblock In \emph{Proceedings of the AAAI Conference on Artificial Intelligence}, pages 10790--10797, 2021.

\bibitem[Yuan et~al.(2025{\natexlab{a}})Yuan, Qu, Qian, Chen, Tang, Sun, Chu, Zhang, Wang, Cai, et~al.]{yuan2025video}
Zhenlong Yuan, Xiangyan Qu, Chengxuan Qian, Rui Chen, Jing Tang, Lei Sun, Xiangxiang Chu, Dapeng Zhang, Yiwei Wang, Yujun Cai, et~al.
\newblock Video-star: Reinforcing open-vocabulary action recognition with tools.
\newblock \emph{arXiv preprint arXiv:2510.08480}, 2025{\natexlab{a}}.

\bibitem[Yuan et~al.(2025{\natexlab{b}})Yuan, Tang, Luo, Chen, Qian, Sun, Chu, Cai, Zhang, and Li]{yuan2025autodrive}
Zhenlong Yuan, Jing Tang, Jinguo Luo, Rui Chen, Chengxuan Qian, Lei Sun, Xiangxiang Chu, Yujun Cai, Dapeng Zhang, and Shuo Li.
\newblock Autodrive-r2: Incentivizing reasoning and self-reflection capacity for vla model in autonomous driving.
\newblock \emph{arXiv preprint arXiv:2509.01944}, 2025{\natexlab{b}}.

\bibitem[Zadeh et~al.(2016)Zadeh, Zellers, Pincus, and Morency]{zadeh2016multimodal}
Amir Zadeh, Rowan Zellers, Eli Pincus, and Louis-Philippe Morency.
\newblock Multimodal sentiment intensity analysis in videos: Facial gestures and verbal messages.
\newblock \emph{IEEE Intelligent Systems}, 31\penalty0 (6):\penalty0 82--88, 2016.

\bibitem[Zadeh et~al.(2018)Zadeh, Liang, Poria, Cambria, and Morency]{zadeh2018mosei}
AmirAli~Bagher Zadeh, Paul~Pu Liang, Soujanya Poria, Erik Cambria, and Louis-Philippe Morency.
\newblock Multimodal language analysis in the wild: Cmu-mosei dataset and interpretable dynamic fusion graph.
\newblock In \emph{Proceedings of the 56th Annual Meeting of the Association for Computational Linguistics (Volume 1: Long Papers)}, pages 2236--2246, 2018.

\bibitem[Zhang et~al.(2025)Zhang, Sun, Hu, Wu, Yuan, Zhou, Shen, and Zhou]{zhang2025pure}
Dapeng Zhang, Jin Sun, Chenghui Hu, Xiaoyan Wu, Zhenlong Yuan, Rui Zhou, Fei Shen, and Qingguo Zhou.
\newblock Pure vision language action (vla) models: A comprehensive survey.
\newblock \emph{arXiv preprint arXiv:2509.19012}, 2025.

\bibitem[Zhang et~al.(2023{\natexlab{a}})Zhang, Wu, Zhang, Hu, Fu, Zhou, and Peng]{zhang2023QMF}
Qingyang Zhang, Haitao Wu, Changqing Zhang, Qinghua Hu, Huazhu Fu, Joey~Tianyi Zhou, and Xi Peng.
\newblock Provable dynamic fusion for low-quality multimodal data.
\newblock In \emph{International conference on machine learning}, pages 41753--41769. PMLR, 2023{\natexlab{a}}.

\bibitem[Zhang et~al.(2023{\natexlab{b}})Zhang, Wu, Zhang, Hu, Fu, Zhou, and Peng]{zhang2023provable}
Qingyang Zhang, Haitao Wu, Changqing Zhang, Qinghua Hu, Huazhu Fu, Joey~Tianyi Zhou, and Xi Peng.
\newblock Provable dynamic fusion for low-quality multimodal data.
\newblock In \emph{International conference on machine learning}, pages 41753--41769. PMLR, 2023{\natexlab{b}}.

\bibitem[Zhang et~al.(2024{\natexlab{a}})Zhang, Wei, Han, Fu, Peng, Deng, Hu, Xu, Wen, Hu, and Changqing]{zhang2024multimodal}
Qingyang Zhang, Yake Wei, Zongbo Han, Huazhu Fu, Xi Peng, Cheng Deng, Qinghua Hu, Cai Xu, Jie Wen, Di Hu, and Zhang Changqing.
\newblock Multimodal fusion on low-quality data: A comprehensive survey.
\newblock \emph{arXiv preprint arXiv:2404.18947}, 2024{\natexlab{a}}.

\bibitem[Zhang et~al.(2024{\natexlab{b}})Zhang, Yoon, Bansal, and Yao]{zhang2024MLA}
Xiaohui Zhang, Jaehong Yoon, Mohit Bansal, and Huaxiu Yao.
\newblock Multimodal representation learning by alternating unimodal adaptation.
\newblock In \emph{Proceedings of the IEEE/CVF Conference on Computer Vision and Pattern Recognition}, pages 27456--27466, 2024{\natexlab{b}}.

\bibitem[Zhou et~al.(2023)Zhou, Wang, Chen, Duan, and Zhu]{zhou2023intra}
Yuwei Zhou, Xin Wang, Hong Chen, Xuguang Duan, and Wenwu Zhu.
\newblock Intra- and inter-modal curriculum for multimodal learning.
\newblock In \emph{Proceedings of the 31st ACM International Conference on Multimedia}, pages 3724--3735, 2023.

\end{thebibliography}
